\pdfoutput=1
\documentclass[11pt]{article}

\usepackage{acl}

\usepackage{times}
\usepackage{latexsym}

\usepackage[T1]{fontenc}
\usepackage[utf8]{inputenc}

\usepackage{microtype}

\usepackage{CJKutf8}

\usepackage{xspace}

\newcommand{\hide}[1]{} %hide

\usepackage{hyperref}
\usepackage{array}
\usepackage{multirow}
\usepackage{makecell}
\usepackage{amsmath}
\usepackage{amssymb}
\usepackage{verbatim}
\usepackage{amsfonts}
\usepackage{graphicx}
\usepackage{color}
\usepackage{float}
\usepackage{bm}
\usepackage{arydshln}
\usepackage[noend]{algpseudocode}
\usepackage{tikz}
\usepackage{booktabs}
\usepackage[linesnumbered, ruled, vlined]{algorithm2e}
\usepackage{threeparttable}
\usepackage{lipsum}

\usepackage{xcolor}
\usepackage{parskip}
\usepackage{longtable}

\newcounter{myenumi}
\newcounter{myenumii}[myenumi]
\title{Inverse is Better! Fast and Accurate Prompt for Few-shot Slot Tagging}
\author{
	Yutai Hou\footnotemark[1]~,
	Cheng Chen\footnotemark[1]~,
	Xianzhen Luo,
	Bohan Li,
	Wanxiang Che\footnotemark[2]~
	\\
	Research Center for Social Computing and Information Retrieval, \\ Harbin Institute of Technology \\
	{ \{ythou, cchen, xzluo, bhli, car\}@ir.hit.edu.cn}  \\
}

\begin{document}
\maketitle
\renewcommand{\thefootnote}{\fnsymbol{footnote}}
\footnotetext[1]{Equal contributions.}
\footnotetext[2]{Corresponding author.}
\renewcommand{\thefootnote}{\arabic{footnote}}

\begin{abstract}
Prompting methods recently achieve impressive success in few-shot learning.
% These methods modify samples with textual prompts and map samples to corresponding labels.
% These methods modify samples with textual prompts and use modified samples as context to decode label tokens, by which they map samples to corresponding labels.
% These methods use samples modified with prompts as context to decode label-related tokens, by which they map samples to corresponding label.
These methods modify input samples with prompt sentence pieces, and decode label tokens to map samples to corresponding labels.
However, such a paradigm is very inefficient for the task of slot tagging. 
Since slot tagging samples are multiple consecutive words in a sentence, the prompting methods have to enumerate all n-grams token spans to find all the possible slots, 
which greatly slows down the prediction. 
To tackle this, we introduce an inverse paradigm for prompting.
Different from the classic prompts mapping tokens to labels, we reversely predict slot values given slot types. Such inverse prompting only requires a one-turn prediction for each slot type and greatly speeds up the prediction.
Besides, we propose a novel Iterative Prediction Strategy, from which the model learns to refine predictions by considering the relations between different slot types.
We find, somewhat surprisingly, the proposed method not only predicts faster but also significantly improves the effect (improve over $6.1$ F1-scores on 10-shot setting) and achieves new state-of-the-art performance.
\end{abstract}

% === submission version ===
% \begin{abstract}
% Prompting methods recently achieve impressive success in few-shot learning.
% These methods embed input samples with prompt sentence pieces and decode label-related tokens to map samples to the label.
% However, such a paradigm is very inefficient for the task of slot tagging. 
% Since the slot tagging samples are multiple consecutive words in a sentence, the prompting methods have to enumerate all n-grams token spans to find all the possible slots, 
% which greatly slows down the prediction. 
% To tackle this, we introduce an inverse paradigm for prompting.
% Different from the classic prompts map tokens to labels, we reversely predict slot values given slot types. Such inverse prompting only requires a one-turn prediction for each slot type and greatly speeds up the prediction.
% Besides, we propose a novel Iterative Prediction Strategy, from which the model learns to refine predictions by considering the relations between different slot types.
% We find, somewhat surprisingly, the proposed method not only predicts faster but also significantly improves the effect (improve over $6.1$ F1-scores on 10-shot setting) and achieves new state-of-the-art performance.
% \end{abstract}

\section{Introduction}

Few-shot learning (FSL) aims at learning a model from only a few examples and is regarded as one of the key steps toward more human-like artificial intelligence \cite{fewshotSurvey}.
Recently, prompt-based methods achieve impressive results and show promising prospects for few-shot learning of Natural Language Processing (NLP) \cite{liu2021makes,zhao2021calibrate}.

Prompt-based methods reformulate a target task into the language modeling problem, which takes advantages of the powerful pretrained Language Models (LM)~\cite{devlin-etal-2019-bert,liu2019roberta,lewis-etal-2020-bart,NEURIPS2020_1457c0d6}. 
For example, when classifying the sentiment of the movie review ``\textit{no reason to watch}'', 
prompting methods insert a piece of text ``\textit{It was}'', i.e. prompts, to the input example, getting ``No reason to watch. It was \_\_''. It is natural to expect a higher probability from the LM to fill the template with ``terrible'' than ``great'', and the original task is then converted to a language modeling task. Such conversion reduces the gap between pretraining and target tasks, which allows less dependency on target task data and helps to achieve better performance in low data scenarios \cite{gao-etal-2021-making}.

\begin{figure}
    \centering
    \includegraphics[width=1.0\linewidth]{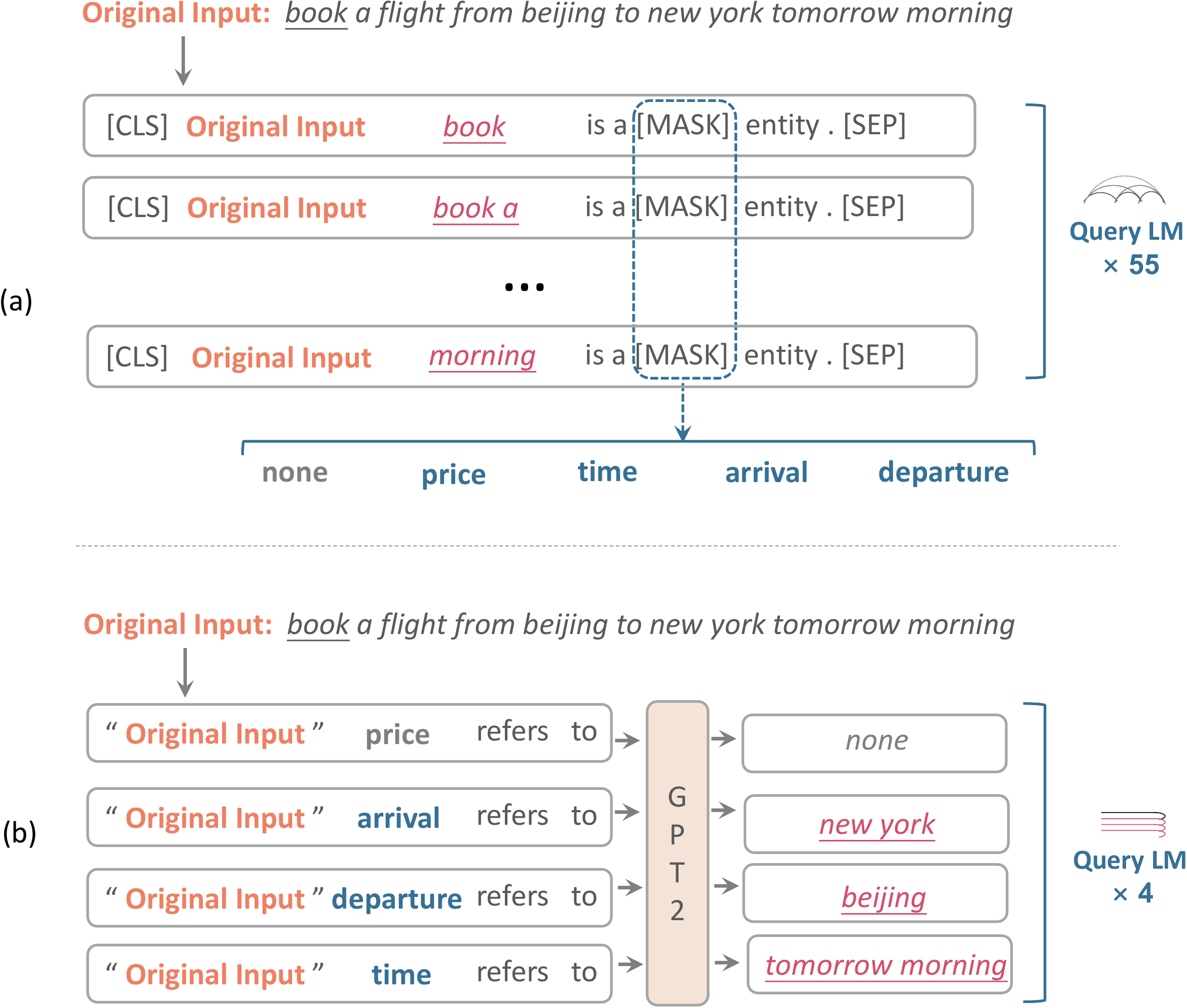}
    \vspace*{-3mm}
    \caption{
    An example of normal (a) and inverse (b) prompting methods for slot tagging.
    For normal prompts, identifying all slots in the query sentence requires enumeration of all spans, while inverse prompt only needs 1-time prediction for each label.  % template-based\wxcomment{前面没提什么是template-based} prompt method. \wxcomment{还需要给出我们方法的图，形成对比}
    }
    \label{fig:intro}
    \vspace*{-3mm}
    % \vspace{-0.2cm}
\end{figure}

However, while achieving great success in sentence level tasks, prompting-based methods show incompatibility for sequence labeling tasks, such as slot tagging.
% and named entity recognition. 
Firstly, the aforementioned prompting paradigm is quite inefficient for slot tagging tasks.
Different from the sentence-level tasks that classify samples of whole sentences, slot tagging samples are multiple consecutive words in a sentence. 
Therefore, as shown in Fig.  \ref{fig:intro}, to find all the possible slots, prompt-based methods have to enumerate all n-gram word spans, and then query LM for each of them, which greatly slows down the prediction \cite{cui-etal-2021-template}.
Further, as a structure prediction problem, slot tagging benefits from taking the
dependencies between labels into account~\cite{ma-hovy-2016-end,Hou-Split}.
For example in Fig.  \ref{fig:intro}, where the \texttt{arrival} entity often appears after a \texttt{departure} entity. 
Such label dependency is hard to be captured by current prompting methods since they predict labels one-by-one independently.

To tackle the above issues, we introduce an inverse paradigm for prompting.
Different from the classic prompts mapping tokens to labels, we reversely predict slot values given slot types.
For the example in Fig.  \ref{fig:intro}, we use an inverse prompt to modify the input as ``\textit{book a flight from Beijing to New York tomorrow morning}. \underline{arrival} \textit{refers to} \_\_'', and then LM is able to decode multi-word span ``New York'' at a time. 
% For the example in Fig.  \ref{fig:intro}, we modify the input with an inverse prompt as ``\textit{book a flight from Beijing to New York tomorrow morning}. \underline{arrival} \textit{refers to} \_\_'', and then LM is able to decode multi-word span ``New York'' at a time. 
Compared to the classic prompts that require predictions for every n-gram word span (55-times in Fig.  \ref{fig:intro}), we only need to perform decoding for $V$-times, where $V$ is the number of label types (4-times in Fig.  \ref{fig:intro}), 
% \wxcomment{比较n-gram的数目和V的大小}
which therefore greatly speeds up the prediction.
Surprisingly, experiments show the proposed method not only predicts faster but also significantly improves the performance, indicating that prompting LM reversely is a better fit for the slot tagging task.
% , which due to that identifying slot values in a given text is 
Besides, to further improve the prediction accuracy, we propose a novel Iterative Prediction Strategy, from which the model learns to refine predictions by considering the relations between different slot types.
% and achieves new state-of-the-art performance
% While label-guided generation obtain comparable results in other task \citation{}, the results indicate  maybe a specific dose for prompting LM to perform seq-labeling task.
% \atmasay{For prompting slot-tagging, inverse paradigm is easier to learn than normal prompting order.}.

To summarize the contribution of this work:
% \begin{itemize}
% \setlength{\itemindent}{0em}
% \setlength{\itemsep}{0em}
% \setlength{\topsep}{-0.5em}
    % \item We introduce the idea of inverse prediction to prompting-methods for slot tagging task, which greatly speeds up the prediction process.
    % \item We propose an iterative strategy for learning and prediction for slot tagging prompt, which allows prompting model to refine prediction and further improve performance.
    % \item Experimental results verify the effectiveness of the proposed method under few-shot setting. 
% \end{itemize}

(1) We introduce the idea of inverse prediction to prompting methods for slot tagging tasks, which greatly speeds up the prediction process.

(2) We propose an Iterative Prediction Strategy for learning and prediction with slot tagging prompts, which allows the prompting model to consider dependency between different slot types and refine prediction.
% and further improve performance.

(3) We extensively evaluate the proposed method in various few-shot settings, where the proposed method brings significant improvements not only in speed but also in accuracy.

The code and data are available at \url{https://github.com/AtmaHou/PromptSlotTagging}.
%\clearpage
%\clearpage
\section{Background}

% \begin{figure*}[t]
% \begin{center}
% \includegraphics[width=1.0\textwidth]{figs/traditional_models.pdf}
% % \vspace*{-3mm}
% \end{center}
% \caption{}
% \label{}
% % \vspace*{-3mm}
% \end{figure*}

\begin{figure*}[t]
\centering
\includegraphics[width=1.0 \textwidth, trim={0.5cm 1.0cm 0.5cm 1.0cm}, clip]{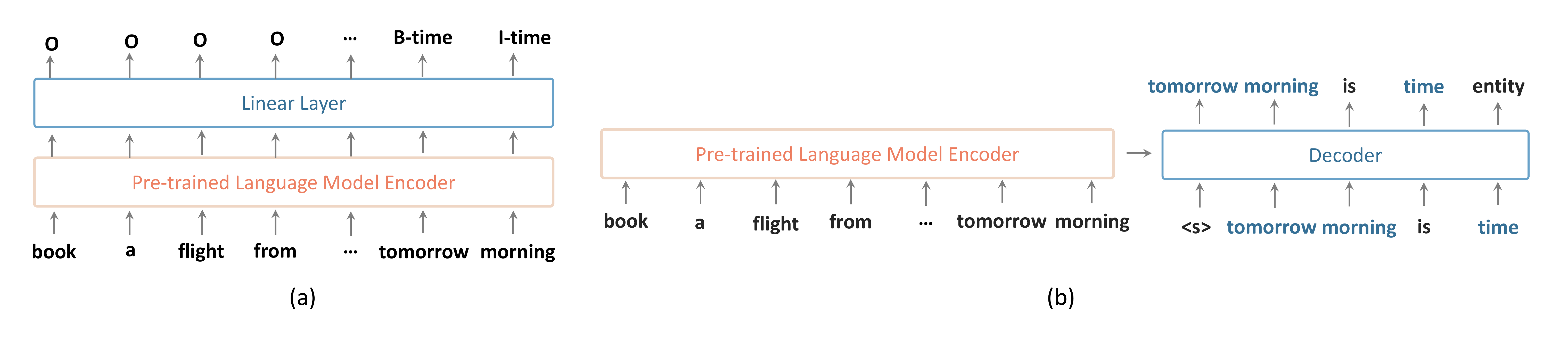}
\caption{
Illustration of conventional sequence labeling method (a) and classic prompting methods (b).
}\label{tranditional_model}
\end{figure*}

In this section, we begin with a formal definition of the few-shot slot tagging task (\S \ref{2.1}), and then introduce the conventional sequence labeling approaches (\S \ref{2.2}) and recent prompts-based methods (\S \ref{2.3}) for this task.

\subsection{Few Shot Slot Tagging}
\label{2.1}
Slot tagging aims at finding key slots within a sentence, such as time or location entities. 
Given an input sentence $\bm{x} = (x_1,x_2,\dots, x_n)$ as a sequence of words,  a slot tagging model extracts all $M$ slot label-values pairs $\bm{y} = \{ (l_i, s_i)\}_{i=1}^M$ in the sentence, where $l_i$ is the $i$th label in the label set $L$ and $s_j^k =\{x_j,...,x_k\} $ is a word span starting from $x_j$ and ending with $x_k$. 

In few-shot settings, model are often evaluated on multiple low-resource domains $\{D_L^{(1)},D_L^{(2)},...\}$, which is called target domain \cite{fewshotSurvey}. 
Each target domain  $D_L^{(j)}$ only contains a few labeled instances called support set $S = \{(\bm{x}^{(i)},\bm{y}^{(i)})\}_{i=1}^{N_S}$, which usually includes \textit{K} examples (K-shot) for each of \textit{N} labels (N-way).
On each target domain, given support set examples as references, few-shot slot tagging models are required to make predictions for query set samples.
Optionally, some few-shot settings also include a set of data-rich domains $\{D_H^{(1)},D_H^{(2)},...\}$ called source domains, which are used for pretraining of few-shot models.

\subsection{Conventional Sequence Labeling Approaches}
\label{2.2}
Conventional approaches often formulate slot tagging as a sequence labeling problem, where each word in input is associated with a sequence label.
Given sentence $\bm{x} = (x_1,x_2,\dots,x_n)$ as input, these method predicts the best-match sequence labels $\bm{y}=(y_1,y_2,...,y_n)$.
To predict slots with multiple words, sequence labeling approaches adopt a ``BIO'' labeling strategy, which uses ``B'' to mark the begin word of a slot, ``I'' to mark the inner words of a slot and ``O'' to mark non-slot words.
For the example in the Fig.  \ref{tranditional_model}, \texttt{B-time} is tagged to the first word in a time slot, \texttt{I-time} is tagged to a non-begin word within a time slot, and \texttt{O} label refers to non-slot words.
As shown in Fig.  \ref{tranditional_model}(a), few-shot sequence labeling model is usually formulated as:
\begin{gather*}
    \bm{h_{1:n}}={\rm Encoder}(\bm{x_{1:n}}), \\
    p(\bm{y_i}|\bm{x}, S)=\rm{Softmax}(\rm{Decoder}(\bm{h_i})), \\
    (i \in [1,2,...,n]), \\
    \bm{y}^* = (y_1, y_2, ..., y_n) = \mathop{\arg\max}\limits_{\bm{y}} p(\bm{y} | \bm{x}, S), 
\end{gather*}
where $S$ is a K-shot support set, $\rm{Encoder}$ is usually a pretrained language model such as BERT~\cite{devlin-etal-2019-bert}, $\bm{h}_{1:n}$ is the hidden state of the encoder with a dimension $d_h$, and $\rm{Decoder}$ can either be a linear layer, a CRF layer or any other parametric or non-parametric classifier.

\subsection{Sequence Labeling with Prompts}
\label{2.3}
Prompt-based methods have been proven effective in many NLU tasks, especially in few-shot settings, but things become complicated when it comes to slot tagging tasks. 
To identify the slot label for a word span $s_i^j =\{x_i,...,x_j\}$ in sentence $\bm{x}$, previous works construct templates, e.g., \textit{“[$\bm{x}$] [$s^j_i$] is a [z] entity”}, and prompt a pretrained language model with such templates to predict label-related words \textit{[z]}~\cite{cui-etal-2021-template}.
For example in the  Fig.  \ref{tranditional_model}(b), predicting the time slot can be achieved as ``\textit{book a flight from Beijing to New York tomorrow morning}. \textit{tomorrow morning is a} \underline{time} \textit{entity.}''
However, to find all possible slots, these methods need to traverse all the n-gram spans $s_i^j ,i,j \in [1,n]$ in a sentence, which is quite expensive in time and computation.

\section{Method}

\begin{figure*}[t]
\begin{center}
\includegraphics[width=1.0\textwidth]{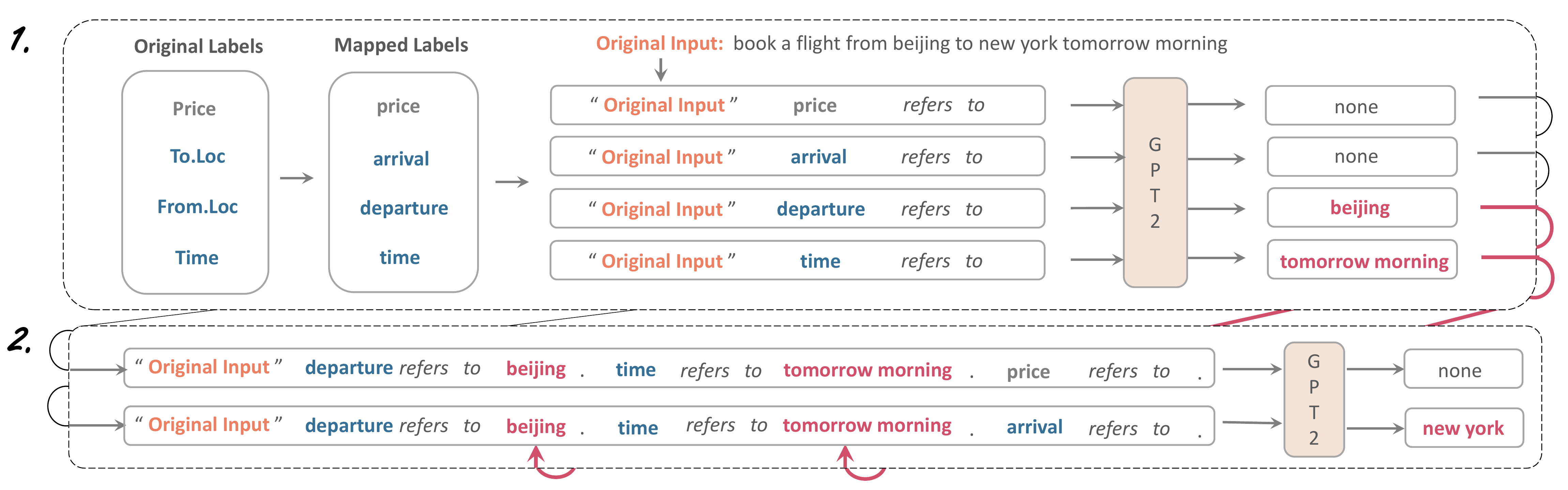}
\end{center}
\caption{Overview of the proposed method with Inverse Prediction and Iterative Prediction Strategy. We first embed the input sentence with inverse prompts and directly decode slot values given slot types. Then we iteratively refine predictions by reinforcing the prompts with predicted slot-value pairs.}
\label{fig:proposed_model}
\end{figure*}

To remedy the high cost of prompt prediction mentioned in the previous section, we introduce a novel inverse paradigm for prompting of slot tagging task, which significantly improves the speed of prediction by transforming the past fill-in-the-blank problem into a generative task.
Specifically, we first introduce the construction of our inverse prompts templates (\S \ref{3.1}), and then describe how to use inverse prompts during training and inference (\S \ref{3.2}). Further, we propose an Iterative Prediction Strategy to refine prediction by considering the relation between different slot types (\S \ref{3.3}).
The overview of proposed method is shown in Fig. \ref{fig:proposed_model}. 

\subsection{Prompt Creation}\label{3.1}
In this section, we introduce the creation of the proposed inverse prompts, which includes three main components: the label mapping, the inverse template and the control tokens. 

\paragraph{Label Mapping}
Before prompt construction, we first need to convert each label into a word form that can be easily understood by the pre-trained language model.
We employ a label mapping process to achieve this, which use a one-to-one mapping function to convert the label set $L = \{l_1, \dots, l_{|L|}\}$ to a natural language word set $\hat{L} =\{\hat{l_1}, \dots , \hat{l_{|L|}}\}$. For example, in Fig.  \ref{fig:proposed_model}, we convert the label set L = $\{$\texttt{from.Loc}, \texttt{to.Loc}, \texttt{Time}, \texttt{Price}$ \}$ to a natural language label set $\hat{L}$ = $\{$ \texttt{departure}, \texttt{arrival}, \texttt{time}, \texttt{price}$ \}$.

\paragraph{Inverse Template}
Prompt template is a piece of sentence with blanks, which is used to modify the original inputs and get prompting inputs for a pretrained language model.
To achieve inverse prompting, our template fills in an original sentence and a label as prefixes and subsequently leaves blanks for the LM to generate the corresponding slot values.
Specifically, given an input sentence $s$ and a set of mapped labels $\hat{L}$, for each mapped label $\hat{l_i}\in \hat{L}$, the inverse template is defined as:
\[
% \vspace*{-1mm}
\text{``$\bm{x}$''  $\hat{l_i}$  \textit{refers  to} \_\_}  
%  \textit{[; \_\_ ...].} 
% \vspace*{-1mm}
\]
For instance, in Fig.  \ref{fig:proposed_model}, we fill the input ``\textit{book a flight from beijing to new york tomorrow morning}'' and each label in $\hat{L}$ into the template to get four prompted inputs $\bm{p}$:\\
\textit{``book a flight from beijing to new york tomorrow morning'' \textbf{departure} refers to \_\_\\
``book a flight from beijing to new york tomorrow morning'' \textbf{arrival} refers to \_\_\\
``book a flight from beijing to new york tomorrow morning'' \textbf{time} refers to \_\_\\
``book a flight from beijing to new york tomorrow morning'' \textbf{price} refers to \_\_}

\paragraph{Control tokens}
Additionally, we introduce control tokens $C$ to complete the prompts function for the slot tagging task.
In order to recognize the case that there's no corresponding entity of the queried slot type, we introduce \texttt{<NONE>} token to pad the output, and in practice, we use \textit{``none''} as \texttt{<NONE>} token to make the model output more natural.
In order to tag more than one entity of the same slot type, we introduce \textit{``;''} as \texttt{<SEP>} to divide more than one entity of the same slot type. 
And we also use \textit{``.''} as \texttt{<END>} token to indicate the end of a single generation.

\subsection{Training and Inference with Inverse Prompts}
\label{3.2}
Till now, we have presented the construction of the inverse prompt. This section will show how to perform training and inference with the prompts.

\paragraph{Training}
At the training time, we pre-construct the prompt with answers such as \textit{``book a flight from beijing to new york tomorrow morning'' \textbf{departure} refers to \textit{\textbf{new york}} .} Then we finetune a pre-trained language model with the answered prompts, and we only calculate loss on the answer tokens (i.e. \textbf{new york}) instead of the loss on the whole sentence.
\begin{gather*}
    L=\sum_{i>|\bm{p}|}\rm{CE}(\hat{y_i},y_i)
\end{gather*}
where $|\bm{p}|$ is the length of the prompted input, $\hat{y}_i$ denotes the model predictions, and ${y}_i$ is the pre-constructed answer.

\paragraph{Inference}
At the inference time, we feed the prompted inputs into the fine-tuned pre-trained language model and let LM generate the appeared slot values. 
During generation, we restrict LM to generate only words that appear in the original input sentence or predefined control words.
For each prompted input $\bm{p}$, the next token $\bm{t}_k \in  \bm{x} \cup  C$ is determined by language model probability:
\[
\vspace*{-1mm}
\bm{t}_k = \mathop{\arg\max}\limits_{\bm{t}_k \in \bm{s} \cup  \bm{C}} \bm{p_\text{LM}}(\bm{t}_k|\bm{p};\bm{t}_{1:k-1})
\vspace*{-1mm}
\]

Note that restricting the scope of output tokens is crucial to the performance.

\subsection{Iterative Prediction Strategy}
\label{3.3}
In the previous section, different slot types are predicted separately. To consider the relations between different slot types, we introduce the Iterative Prediction Strategy, which also provides the model a second chance to revise those unrecognized entities.
We assume that different labels are interactive, so the predicted slots could be used as a hint to help predict the missed slots. 
For example in Fig. \ref{fig:proposed_model}, it is often easier to generate the ``arrival'' slot given the results of ``departure'' and ``time''.
Motivated by this, as shown in the Fig. \ref{fig:proposed_model}, we construct another template that concatenates those filled prompts as additional generation condition and use them to revise the slot values that are ``\textit{\textbf{none}}'' in the first round of prediction. 
Below we describe the details of this strategy during training and testing.

\paragraph{Training}
At the training time, we simulate the cases where the slots are not recognized to enable the model to revise the \textit{\textbf{none}} slot values. We do this by randomly constructing \textit{\textbf{none}} slot value examples. For example, at training time, suppose there are four training prompts filled with true answers:\\
\textit{``book a flight from beijing to new york tomorrow morning'' \textbf{departure} refers to \textbf{beijing .}\\
``book a flight from beijing to new york tomorrow morning'' \textbf{arrival} refers to \textbf{new york .}\\
``book a flight from beijing to new york tomorrow morning'' \textbf{time} refers to \textbf{tomorrow morning .}\\
``book a flight from beijing to new york tomorrow morning'' \textbf{price} refers to \textbf{none .}}\\
We randomly select some occurred labels (e.g., ``arrival'') pretending it was not predicted, and construct a second round prompt:\\
\textit{``book a flight from beijing to new york tomorrow morning'' \textbf{departure} refers to \textbf{beijing .} \textbf{time} refers to \textbf{tomorrow morning .} \textbf{price} refers to \textbf{none .} \textbf{arrival} refers to \_\_}.\\
By using these second round prompts for model training, we encourage the language model to find those unrecognized slots in the first round prediction and allow the model to consider relationships between labels.

\paragraph{Inference}
During the inference time, we construct the second-round prompts and revise the slots that are not recognized in the first round. 
For example in the Fig.  \ref{fig:proposed_model}, the model predict \textit{\textbf{none}} value for ``price'' and ``arrival'' slot in the first round. 
We then construct another iteration of the prompted inputs that query the unrecognized slots, given all the labels and slot values that have been predicted:\\
\textit{``book a flight from beijing to new york tomorrow morning'' \textbf{departure} refers to \textbf{beijing .} \textbf{time} refers to \textbf{tomorrow morning .}  \textbf{arrival} refers to \_\_}.\\
\textit{``book a flight from beijing to new york tomorrow morning'' \textbf{departure} refers to \textbf{beijing .} \textbf{time} refers to \textbf{tomorrow morning .}  \textbf{price} refers to \textbf{\_\_ .}}

\noindent The model is expected to predict the first-round missed slots during the second iteration, considering relations between labels.
%\clearpage
%\input{3_problem}
\section{Experiment}
We evaluate the performance of the proposed method on two classic few-shot scenarios: ($1$) Setting with Only In-domain data, where all training data are only a few labeled support data.
($2$) Setting with Meta Source Tasks, where some additional data-rich source domains are available for pretraining.

\paragraph{Evaluation}
To use same evaluation criteria as conventional sequence labeling methods, we need to label tokens reversely and get output in same format. 
After generation, we first separate outputs into slot values. For each slot value, we label tokens in the source sentence with three principles: ($1$) Slot value is complete: only if the whole slot value matches a span in the source sentence, we label it with the corresponding label. ($2$) Choose the first overlap predicted slot span: if any token in the source sentence has been labeled, we do not relabel this token even when it matches another slot value. ($3$) Use BIO labels: add ``B-'' to the beginning token of the slot span, add ``I-'' to the non-begin token of the slot span, and label non-slot tokens with ``O''. 
After labeling tokens reversely, we evaluate F1 scores within each few-shot episode.\footnote{
	For each episode, 
	we calculate the F1 score on query samples with \texttt{conlleval} script:  \url{https://www.clips.uantwerpen.be/conll2000/chunking/conlleval.txt}
}

\begin{table*}[t]
	\centering
	\footnotesize
	\begin{tabular}{lccccccc}
	\toprule
		\multirow{2}{*}{\textbf{Model}} &
		\multicolumn{6}{c}{\textbf{MIT-Restaurant}} & \\
		\cmidrule(lr){2-7}
		
		& {\textbf{10}} & {\textbf{20}} & {\textbf{50}} & {\textbf{100}} & {\textbf{200}} & {\textbf{500}} \\
		\midrule
       ~\citet{wiseman-stratos-2019-label} + PT   & 4.1 & 3.6 & 4.0 & 4.6 & 5.5 & 8.1 \\
       ~\citet{Ziyadi-2020} + PT   & 27.6 & 29.5 & 31.2 & 33.7 & 34.5 & 34.6 \\
       ~\citet{new5huang2020few} + PT   & 46.1 & 48.2 & 49.6 & 50.0 & 50.1 & - \\
        {Sequence Labeling BART} + PT   & 8.8 & 11.1 & 42.7 & 45.3 & 47.8 & 58.2 \\
        {Sequence Labeling BERT} + PT  & 27.2 & 40.9 & 56.3 & 57.4 & 58.6 & 75.3 \\
        {Template-based BART} + PT   & \textbf{53.1} & 60.3 & 64.1 & 67.3 & 72.2 & 75.7 \\
        \hdashline
        {Sequence Labeling BERT}  & 21.8 & 39.4 & 52.7 & 53.5 & 57.4 & 61.3 \\
        {Template-based BART}  & 46.0 & 57.1 & 58.7 & 60.1 & 62.8 & 65.0 \\
        {Ours}   & 49.35 & 60.48 & 65.34 &  70.41 &  73.69 & 76.13 \\
		{Ours + Iterative}   & {\textbf{52.10}} & {\textbf{61.49}} &  {\textbf{66.83}} &  {\textbf{70.98}} &  {\textbf{73.97}} & {\textbf{76.37}} \\
	    \toprule
		\multirow{2}{*}{\textbf{Model}} &
		\multicolumn{6}{c}{\textbf{MIT-Movie-Hard}} & \\
		\cmidrule(lr){2-7}
		
		& {\textbf{10}} & {\textbf{20}} & {\textbf{50}} & {\textbf{100}} & {\textbf{200}} & {\textbf{500}} \\
		\midrule
       ~\citet{wiseman-stratos-2019-label} + PT   & 3.1 & 4.5 & 4.1 & 5.3 & 5.4 & 8.6 \\
       ~\citet{Ziyadi-2020} + PT   & 40.1 & 39.5 & 40.2 & 40.0 & 40.0 & 39.5 \\
       ~\citet{new5huang2020few} + PT   & 36.4 & 36.8 & 38.0 & 38.2 & 35.4 & 38.3 \\
        {Sequence Labeling BART} + PT   & 13.6 & 30.4 & 47.8 & 49.1 & 55.8 & 66.9 \\
        {Sequence Labeling BERT} + PT  & 28.3 & 45.2 & 50.0 & 52.4 & 60.7 & 76.8 \\
        {Template-based BART} + PT   & 42.4 & 54.2 & 59.6 & 65.3 & 69.6 & {\textbf{80.3}} \\
        \hdashline
        {Sequence Labeling BERT}  & 25.2 & 42.2 & 49.64 & 50.7 & 59.3 & 74.4 \\
        {Template-based BART}  & 37.3 & 48.5 & 52.2 & 56.3 & 62.0 & 74.9 \\
        {Ours}   & 52.07 & 59.11 & 65.63 &  69.35 & 72.36 & {\textbf{75.03}}  \\
		{Ours + Iterative}   & {\textbf{53.31}} & {\textbf{60.19}} &  {\textbf{66.13}} &  {\textbf{69.63}} & {\textbf{72.45}} & 74.83 \\
		\toprule
		\multirow{2}{*}{\textbf{Model}} &
		\multicolumn{6}{c}{\textbf{MIT-Movie}} & \\
		\cmidrule(lr){2-7}
		
		& {\textbf{10}} & {\textbf{20}} & {\textbf{50}} & {\textbf{100}} & {\textbf{200}} & {\textbf{500}} \\
		\midrule
        {Sequence Labeling BERT}   & 50.60 & 59.34 & 71.33 & - & - & - & \\
        {NNShot}   & 50.47 & 58.94 & 71.17 & - & - & - & \\
        {StructShot}  & 53.19 & 61.42 & 72.07 & - & - & - & \\
        {Template-based BART}   & 49.30 & 59.09 & 65.13 & - & - & - & \\
        {EntLM}  & 57.31 & 62.36 & 71.93 & - & - & - & \\
        {Ours}   & 57.04 & 67.86 &  76.81 &  80.28 &  82.43 & {\textbf{84.55}} \\
		{Ours + Iterative}   & {\textbf{59.74}} & {\textbf{70.09}} &  {\textbf{77.60}} &  {\textbf{80.63}} &  {\textbf{82.64}} & 84.51 \\
		\bottomrule
	\end{tabular}
	\caption{ \footnotesize  
		F1 scores of few-shot slot tagging task on three different datasets.10
        indicates 10 instances for each entity type.
    \texttt{+PT} denotes the model is pre-trained on additional datasets.
		\texttt{+Iterative} denotes enhance model with Iterative Prediction Strategy.
	}\label{tbl:mit}
% 	\vspace*{-4mm}
\end{table*}

\subsection{Setting with Only In-domain data}
\paragraph{Datasets}
For few-shot setting without source domain transfer, we conduct experiments on three few-shot datasets with only in-domain data:
MIT-Restaurant Review~\cite{liu2013query}, MIT-Movie  Review~\cite{liu2013query} and MIT-Movie-Hard Review.\footnote{MIT-Movie Review has two datasets: a simple one and a complex one. We denote the simple one as MIT-Movie and combine both as MIT-Movie-Hard.}
We conduct experiments with $K \in \{10,20,50,100,200,500\}$ shots settings to fully evaluate the performance of our method in all three datasets.
To overcome the randomness associated with support set selection, we sample 10 different support set for each $K$-shot setting and report averaged results.
All models are trained and tested with the same data.

\paragraph{Implements}
Our model employs the smallest GPT2~\cite{radford2019language} pre-trained model as the base model for fine-tuning, and no new parameters are introduced. Besides, we set the learning rate as $6.25e-5$ and batch size as $2$ for few-shot training. 
For all our experiments, we finetune the model only on few-shot support set for $2$ epochs ($4$ on $10$/$20$ shots settings)
with the AdamW optimizer and linear decaying scheduler.
Since there is no development set, all hyperparameters are roughly set based on experience without tuning.
Data and code used are public available.

\paragraph{Baselines}
In our experiments, we compare with competitive baselines including both conventional sequence labeling methods and recent prompt-based methods. 

% \noindent $\bullet$
\noindent $\bullet$ \textbf{Sequence Labeling BERT}~\cite{devlin-etal-2019-bert} can be seen as a BERT-based sequence labeling baseline which fine-tunes the BERT model with a token-level linear classifier head. 

\noindent $\bullet$ \textbf{Template-based BART}~\cite{cui-etal-2021-template} %A template-based prompt method using 
is a prompt-based method that query BART-based LM \cite{lewis-etal-2020-bart} every possible span in sentence if it belong to a certain category and therefore also need to enumerate all label for inference.

\noindent $\bullet$ \textbf{NNShot and StructShot}~\cite{yang-katiyar-2020-simple} are two
metric-based few-shot learning approaches for slot tagging and NER. NNShot is an instance-level nearest neighbor classifier for few-shot prediction, and StructShot promotes NNShot with a Viterbi algorithm during decoding.

\noindent $\bullet$ \textbf{EntLM}~\cite{EntLM} is a prompt-based method that leverage substitution between words of the same type to achieve
one pass prediction.

\begin{table*}[t]
	\centering
	\footnotesize
    % \small
	\begin{tabular}{lccccccccc}\toprule
		\multirow{2}{*}{\textbf{Model}} &
		\multicolumn{7}{c}{\textbf{5-shot Slot Tagging}} & \\
		\cmidrule(lr){2-8}
		& {\textbf{We}} & {\textbf{Mu}} & {\textbf{Pl}} & {\textbf{Bo}} & {\textbf{Se}} & {\textbf{Re}} & {\textbf{Cr}} & {\textbf{Ave.}} \\
		\midrule
		{Bi-LSTM}  & 25.44 & 39.69 & 45.36 & 73.58 & 55.03 & 40.30 & 40.49 & 45.70  \\
		{SimBERT }       & 53.46 & 54.13 & 42.81 & 75.54 & 57.10 & 55.30 & 32.38 & 52.96 \\
		{TransferBERT }    & 56.01 & 43.85 & 50.65 & 14.19 & 23.89 & 36.99 & 14.29 & 34.27 \\
        {MN}  & 38.80 & 37.98 & 51.97 & 70.61 & 37.24 & 34.29 & 72.34 & 49.03 \\
		{WPZ+BERT} & 69.06 & 57.97 & 44.44 & 71.97 & 74.62 & 51.01 & 69.22 & 62.61 \\
        {TapNet+CDT}     & 67.83 & 68.72 & 73.74 & 86.94 & 72.12 & 69.19 & 66.54 & 72.15 \\
		{L-WPZ+CDT}        & {\textbf{78.23}} & 62.36 & 59.74 & 76.19 & 83.66 & 69.69 & 71.51 & 71.62 \\
        {L-TapNet+CDT}     & 69.58 & 64.09 & 74.93 & 85.37 & {\textbf{83.76}} & 69.89 & 73.80 & 74.49 \\
        ConVEx* & { 71.5} & { \textbf{77.6} } & {\textbf{ 79.0}} & { 84.5 } & { 84.0 } & {\textbf{ 73.8} } & { 67.4 } & { \textbf{76.8} }\\
		\midrule
        {Ours}   & 70.44 & 71.63 & 78.67 & {\textbf{87.37}} & 81.38 & 71.77 & 74.42 & 76.53 \\
		{Ours + Iterative}   & 70.63 & {\textbf{71.97}} & {\textbf{78.73}} & 87.34 & {81.95} & {\textbf{72.07}} & {\textbf{74.44}} & {\textbf{76.73}} \\
		\bottomrule
	\end{tabular}
	\caption{ \footnotesize 
		F1 score results on 5-shot Snips.
		* denotes using additional Reddit data for pretraining. Our methods achieve the best performance among those using same training data.}\label{tbl:snips}
% 	\vspace*{-4mm}
\end{table*}

\paragraph{Results}
Table \ref{tbl:mit} shows the results of the proposed method only finetuned on few-shot in-domain data. Among these results, we can observe that:

($1$) Our proposed method performs consistently better than all the baseline methods on all three datasets. It outperforms the strongest baseline Template-based BART which uses BART-large by average F1 scores on three datasets of $11.96$ in $10$-shot setting even with a much smaller pre-trained language model (the smallest GPT2).

($2$) Our proposed method is even comparable or outperforms those baselines with data-rich domain pre-training.

($3$) Our proposed method performs much better than baselines in fewer labeled samples settings, especially in $10$ and $20$ shot settings, which indicates our method can leverage information from limited labeled data more efficiently.

($4$) Our method significantly outperformed Sequence Labeling BERT whose performance is quite poor on $10$ and $20$ shot settings, which indicates that the number of labeled data is too scarce for conventional sequence labeling tasks, and proves that the prompt-based method is effective in few-shot slot tagging tasks.

($5$) The proposed Iterative Prediction Strategy consistently improves the slot tagging performance. The improvements become greater with fewer learning shots and the averaged improvement in $10$ and $20$ shot setting on three datasets are $2.23$ and $1.44$. This shows that when there is less data, the iterative revising mechanism is more important.

\subsection{Setting with Meta Source Tasks}
\paragraph{Datasets}
We also evaluate the model ability of transferring from data-rich domains to unseen few-shot domains and conduct experiments on SNIPS~\cite{SNIPS} dataset. 
Following the data split provided by~\citet{Hou-Split}, we construct $5$-shot SNIPS datasets from the original SNIPS datasets.
The few-shot SNIPS dataset consists of $7$ domains with different label sets: GetWeather (We), Music (Mu), PlayList (Pl), RateBook (Bo), SearchScreenEvent (Se), BookRestaurant (Re), and SearchCreativeWork (Cr).
Each domain contains $100$ few-shot episodes, and each episode consists of a support set and a query.

\paragraph{Implements}
Following \citet{ConVEx}, we conduct our cross-domain experiments with $5$-shot few-shot settings to evaluate the ability of our model to transfer from rich-data domains to unseen few-shot domains. For our proposed method, same as in-domain settings, we use the smallest GPT2 as the base model, and no new parameters are introduced. We pretrain the model in source domains and fine-tune it on the target few-shot domain. We set learning rate as $6.25e-5$ and batch size as $16$ for pretraining and batch size as $2$ for $5$-shot finetuning.
During finetuning, we use the same AdamW optimizer and linear decaying scheduler.
The hyper-parameters are decided according to performance on the dev set. 
Data and code used are public available.

\paragraph{Baselines}
We provided competitive strong baselines, including traditional finetune-based methods and advanced few-shot learning methods.

\noindent $\bullet$ \textbf{Bi-LSTM}~\cite{Bi-LSTM} uses GLoVe~\cite{Glove} embedding for slot tagging and is trained on the support sets.

\noindent $\bullet$ \textbf{SimBERT} is a metric-based method using cosine similarity of BERT-based embedding to label tokens with the most similar token's label.

\noindent $\bullet$ \textbf{Matching Network (MN)} \cite{MN} is a few-shot sequence labeling model based on the matching network and uses BERT embedding.

\noindent $\bullet$ \textbf{TransferBERT} is a domain transfer-based conventional NER model using BERT, which is first pre-trained on source domains and then fine-tuned on the target domain support set.

\noindent $\bullet$ \textbf{WPZ}~\cite{Fritzler_2019} is a metric-based few-shot slot tagging method similar to MN, but is based on the prototypical network~\cite{NIPS2017_cb8da676}.

\noindent $\bullet$ \textbf{TapNet+CDT}, \textbf{L-TapNet+CDT}, \textbf{L-WPZ+CDT} \cite{Hou-Split} are metric-based few-shot learning methods designed for slot tagging, which introduces a CRF-based framework to consider the relation between different slots.

\noindent $\bullet$ \textbf{ConVEx}~\cite{ConVEx} is a fine-tuning-based method that models slot tagging as a cloze task and is first pre-trained on Reddit data then fine-tuned on few-shot slot tagging data. Note that the Reddit data is not used by our method and other baselines during the experiments.

\paragraph{Results}
Table~\ref{tbl:snips} shows the results of the cross-domain few-shot setting, from which we can observe that:

($1$) Our proposed method outperforms all the baselines except ConVEx which uses extra Reddit data in the cross-domain 5-shot setting. Despite using less training data, our model still achieves comparable results with Covex, proving its superiority.

($2$) We outperform TransferBERT by $42.36$ F1 scores which strongly proved that the prompt-based method can transfer more knowledge from the source domain and is more data-efficient than conventional methods.
% Noting that we can directly compare with TransferBERT for both our methods first pre-trained on source domains and then finetuned on each few-shot domain respectively without any few-shot learning tricks. 

(3) Our method outperforms metric-based few-shot learning baselines, for example, $2.24$ F1 scores higher than L-TapNet+CDT, which proves its competitiveness compared to classical few-shot learning methods.

(4) Our Iterative Prediction Strategy improved Our method by about $0.5$ F1 scores, demonstrating that the revising ability is likely to be transferable and is effective under cross-domain scenarios.

\subsection{Analysis}
% List analysis experiment results table and analysis here.

\paragraph{Effects of Iterative Prediction Strategy}
As shown in Table \ref{tbl:mit}, the proposed Iterative Prediction Learning brings consistent improvement, 
% to our method, 
especially in low-resource settings. It works by revising predictions with a second-round query to recognize those missing slots, which can bring an increase in recall score. 
To confirm that, we make a detailed analysis with precision score (P), recall score (R) and F1 score (F) in Table \ref{tbl:ablation}. 

When Iterative Revise Strategy is added, we can get a rise in recall score about $4$ points in $10$-shot, $2$\textasciitilde$4$ points in $20$ shot and more than $1$ points in other shot settings in exchange for a slight precision drop, resulting in a rise in overall F1 score by about $2$ points in $10$ and $20$ shots.

We further explore the effect of jointly learning of the first-round prediction and the second-round revising, and learn two abilities separately with two models. As shown in Table \ref{tbl:ablation}, w/o Joint model outperforms the no-revising model but lags behind the joint model. This indicates that joint learning the revising ability may act as data augmentation and brings more improvements than simple revising.

\begin{table}[t]
	\centering
	\footnotesize
    % \small
	\setlength{\tabcolsep}{5pt}
    \resizebox{\linewidth}{!}{
	\begin{tabular}{ll ccc ccc}\toprule
		& \multirow{2}{*}{\textbf{Model}} &
		\multicolumn{3}{c}{\textbf{MIT-Restaurant}} & 
		\multicolumn{3}{c}{\textbf{MIT-Movie}} 
		\\
		\cmidrule(lr){3-5}
		\cmidrule(lr){6-8}
		& & {\textbf{P}} & {\textbf{R}} & {\textbf{F}} 
		& {\textbf{P}} & {\textbf{R}} & {\textbf{F}}
		\\
		\midrule
        \multirow{3}{*}{\textbf{10}} 
        & {Ours}   & 67.7 & {\textbf{42.4}} & {\textbf{52.1}} 
		& 84.0 & {\textbf{46.4}} & {\textbf{59.7}} \\
        & {w/o Iter}   & 69.4 & 38.3 & 49.4 
        & 85.9 & 42.7 & 57.0 \\
		& {w/o Joint}   & 68.8 & 38.9 & 49.7 
        & 85.6 & 43.0 & 57.2 \\
		\midrule
        \multirow{3}{*}{\textbf{20}}  
        
		& {Ours}   & 70.1 & {\textbf{54.7}} & {\textbf{61.5}} 
		& 83.5 & {\textbf{60.4}} & {\textbf{70.1}}\\
		& {w/o Iter}   & 71.6 & 52.3 & 60.5 
		& 86.3 & 55.9 & 67.9\\
		& {w/o Joint}   & 70.92 & 53.45 & 61.0 
		& 85.6 & 56.9 & 68.3\\
		\midrule
		\multirow{2}{*}{\textbf{50}}  
        
		& {Ours}   & 73.6 & {\textbf{61.2}} & {\textbf{66.8}} 
		& 83.6 & {\textbf{72.4}} & {\textbf{77.6}}\\
		& {w/o Iter}   & 75.4 & 57.6 & 65.3
		& 85.9 & 69.5 & 76.8\\
		& {w/o Joint}   & 74.3 & 59.2 & 65.7 
		& 84.7 & 70.8 & 77.1\\
		\midrule
		\multirow{3}{*}{\textbf{100}}  
        
		& {Ours}   & 76.1 & {\textbf{66.5}} & {\textbf{71.0}} 
		& 84.4 & {\textbf{77.2}} & {\textbf{80.6}}\\
		& {w/o Iter}   & 78.0 & 64.2 & 70.4 
		& 86.3 & 75.0 & 80.3\\
		& {w/o Joint}   & 76.7 & 66.0 & 71.0 
		& 85.0 & 76.5 & 80.5\\
		\midrule
		\multirow{3}{*}{\textbf{200}}  
        
		& {Ours}   & 77.8 & {\textbf{70.5}} & {\textbf{74.0}} 
		& 85.4 & {\textbf{80.0}} & {\textbf{82.6}}\\
		& {w/o Iter}   & 79.5 & 68.7 & 73.7 
		& 87.1 & 78.2 & 82.4\\
		& {w/o Joint}   & 78.0 & 70.1 & 73.8 
		& 85.1 & 79.9 & 82.4\\
		\midrule
		\multirow{3}{*}{\textbf{500}}  
        
		& {Ours}   & 79.4 & {\textbf{73.5}} & {\textbf{76.4}} 
		& 86.3 & {\textbf{82.8}} & {84.5}\\
		& {w/o Iter}   & 81.0 & 71.8 & 76.1 
		& 87.9 & 81.4 & \textbf{{84.6}}\\
		& {w/o Joint}   & 79.6 & 73.4 & 76.4 
		& 86.6 & 82.1 & 84.3\\
		
		\bottomrule
	\end{tabular}
	}
	\caption{\footnotesize 
		Ablation analysis Iterative Prediction Strategy
		\textbf{w/o Iter} denotes removing iterative strategy and \textbf{w/o joint} denotes using two separate models for the two iterative steps. 
	}\label{tbl:ablation}
% 	\vspace*{-5mm}
\end{table}

\paragraph{Efficiency Study}
Unlike Template-based BART that queries every n-gram span in the source sentence for each label (with ${O(n^2*m)}$ where $n$ is the length of the source sentence and $m$ is the size of the label set) time complexity, our proposed method queries labels in the label set and directly generate slot values (with ${O(n*m)}$ time complexity).
In theory, our method is much faster than Template-based BART, especially dealing with long sentences with sparse slots. 
To prove this, we conduct efficiency experiments by calculating the decoding time of each method on a Titan XP GPU with batch size as $8$, and we set our max generation length at $40$.
As shown in Table \ref{tbl:time}, our method is about $8$ times as fast as the Template-based BART method, and more than $3$ times as fast as theirs with Iterative Prediction Strategy. 
During experiments, we find that as the number of labels increases, the model does become linearly slower, which may become limitations.
However, the number of label types is usually smaller than the sentence length and much smaller than the number of spans, so that this growth does not affect the value of our method in practice.
Besides, we find no significant correlation between the number of labels and our performance.

\begin{table}[t]
	\centering
	\footnotesize
    % \small
    \setlength\tabcolsep{3pt}
	\begin{tabular}{lcc}\toprule
		\multirow{1}{*}{\textbf{Model}} &
		\multicolumn{1}{c}{\textbf{MIT-Movie}} & 
		\multicolumn{1}{c}{\textbf{MIT-Restaurant}} 
		\\
		\midrule
        {Baseline (Normal Prompt)}  & 408.0 &  236.0  \\
        {Ours}  & 51.2 & 33.2 \\
		{Ours + Iterative}   & 119.4 & 71.4 \\
		\bottomrule
	\end{tabular}
	\caption{ \footnotesize 
		Comparison of the decoding time (second).
	}\label{tbl:time}
%  	\vspace*{-4mm}
\end{table}

%\clearpage
\section{Related Work}

\paragraph{Prompt-based learning}
Prompt-based learning approaches have been a broadly discussed topic since large language models like GPT models~\cite{NEURIPS2020_1457c0d6} are hard to fine-tune in low-resource scenarios. Early attempts \cite{schick-schutze-2021-exploiting,schick-schutze-2021-just} introduce manual prompts to text classification tasks.
For natural language understanding (NLU) tasks, automatically searching discrete prompts methods are proposed such as~\citet{jiang2020can,shin2020autoprompt,gao-etal-2021-making}. Meanwhile, due to the continuity of parameters in neural networks, continuous prompts for both text classification and generation tasks~\cite{li2021prefix,DBLP:journals/corr/abs-2103-10385,han2021ptr} have been proposed.  Unlike sentence-level tasks, prompting methods are very complicated for slot tagging and NER tasks. \citet{cui-etal-2021-template} proposes a template-based method querying every slot span with each label which is expensive for decoding. Different from them, we introduce an inverse paradigm for prompting slot tagging tasks.
% by generate slots, different from 
Note that inverse prompting~\cite{DBLP:conf/kdd/ZouYZYYT21} has a similar name to our work but is entirely different in method and task. They aim to generate prompt templates inversely.
Amendable generation~\cite{DBLP:journals/corr/abs-2110-15659} share a similar idea of using Iterative Prediction Strategy to generate and revise dialog state. By contrast, we focus on a different task for sequence labeling and first introduce an Iterative Prediction Strategy to prompting models.
There are also generation-based methods for sequence labeling \cite{genner2021}, which is not a prompting method, since it re-initializes decoding layers and learns a generative model from scratch.

\paragraph{Few-shot slot tagging}
Previous few-shot slot tagging methods focus on metric learning based methods, which classify tokens by word-label similarity \cite{NIPS2017_cb8da676,MN}.
\citet{Hou-Split} leverage label name semantics to get better label representation and model label dependency in few-shot settings.
\citet{yang-katiyar-2020-simple} make a prediction based on the nearest neighbor sample instead of the nearest label representation. 
Besides, some works also explore training a model with additional data from non-slot-tagging task \cite{new5huang2020few,ConVEx}. 
\citet{hou2021learning} improves few-shot slot tagging performance by jointly learning it with intent detection. 
% For example, \cite{ConVEx} employs cloze task data from Reddit to reinforce slot tagging model training.
Different from directly learning the few-shot slot tagging model, some researches explore to reformulate the slot tagging into other NLP tasks. \citet{new12DBLP:conf/acl/MaYLZ21} reforms slot tagging into a reading comprehension task.  \citet{new13DBLP:conf/naacl/YuHZDPL21} treats slot tagging as a retrieval task, \citet{new14DBLP:conf/acl/CoopeFGVH20} uses span extracting task to extract slot and predict corresponding label and \citet{cui-etal-2021-template} leverages prompts for few-shot NER. Different from those methods above, we are the first to reformulate the slot tagging task into a prompt-based generation task. 

\section{Conclusion}
In this paper, to liberate the prompting methods from the burdensome prediction of slot-tagging tasks, we introduce a novel inverse prediction manner to prompting methods of slot-tagging, which significantly improves both the efficiency and accuracy. 
To further improve performance, we propose an Iterative Prediction Strategy for learning, which enables the prompting model to consider dependency between labels and refine prediction.
Extensive experiments verify the effectiveness of the proposed method in various few-shot settings, indicating inverse prediction is a better fit for prompting of slot tagging task.
% where the proposed method improves not only the speed, but also the prediction accuracy.

\section*{Acknowledgments}
We are grateful for the helpful comments and suggestions from the anonymous reviewers. 
This work was supported by the National Key R\&D Program of China via grant 2020AAA0106501 and the National Natural Science Foundation of China (NSFC) via grant 61976072 and 62176078.

\section*{Ethics Section}
We analyze the limitations of the proposed method in both efficiency and effectiveness aspects, and the proposed method has no obvious potential risks.
All the scientific artifacts used/created are properly cited/licensed, and the usage is consistent with their intended use. 
This paper does not collect new datasets, nor does the data used contain sensitive information.

\bibliography{ref}
\bibliographystyle{acl_natbib}

% \appendix

\end{document}